\documentclass[letterpaper, 10 pt, conference]{ieeeconf}
\IEEEoverridecommandlockouts  

\usepackage{lipsum} 
\usepackage{cite}
\usepackage{float}
\usepackage{graphicx}
\usepackage{dsfont}
\usepackage{amssymb}
\usepackage{multirow}

\usepackage{amsmath} 
\usepackage{mathtools}  
\usepackage{amsfonts}
\usepackage{booktabs}
\usepackage{pgfplots}
\usepackage{siunitx}
\usepackage{enumerate}
\usepackage{tabularx}

\usepackage{nccmath}

\usepackage{amsthm}

\theoremstyle{plain}

\usepackage{float}
\floatstyle{plaintop}
\restylefloat{table}
\usepackage{MnSymbol}%
\usepackage{wasysym}%
\usepackage{caption}
\usepackage{subcaption}
\usepackage{graphicx}
\usepackage{import}
\usepackage{bbm}
\usepackage{textcomp}
\usepackage[hidelinks]{hyperref}

\usepackage[colorinlistoftodos]{todonotes}
\usepackage{algorithm}
\usepackage{algpseudocode}

 \renewcommand{\baselinestretch}{0.96}
\definecolor{deepblue}{rgb}{0,0,1}

\newcommand{\revisions}[1]{{#1}}
\newcommand{\qbreview}[1]{}

\begin{document}

\title{
The Spinning Blimp: Design and Control of a Novel Minimalist Aerial Vehicle Leveraging Rotational Dynamics and Locomotion
}

%

\author{Leonardo Santens$^*$,  Diego S. D'Antonio$^*$, Shuhang Hou, and David Saldaña 
\thanks{${}^*$ The authors contributed equally.}
\thanks{$^1$The authors are with the Autonomous and Intelligent Robotics Laboratory --AIRLab-- at Lehigh University, Bethlehem, PA, 18015, USA. \texttt{Email:\{les422, shh420, diego.s.dantonio, saldana\}@lehigh.edu}}
\thanks{The authors gratefully acknowledge the support of the NSF Award 2322840 and ONR 544835.}
}

\maketitle

\begin{abstract}

This paper presents the Spinning Blimp, a novel lighter-than-air (LTA) aerial vehicle
designed for low-energy stable flight. Using an oblate spheroid helium balloon for buoyancy, the vehicle achieves minimal energy consumption while maintaining prolonged airborne states. The unique and low-cost design employs a passively arranged wing coupled with a propeller to induce a spinning behavior, providing inherent pendulum-like stabilization. 
We propose a control strategy that takes advantage of the continuous revolving nature of the spinning blimp to control translational motion. 
The cost-effectiveness of the vehicle makes it highly suitable for a variety of applications, such as patrolling, localization, air and turbulence monitoring, and domestic surveillance. 
Experimental evaluations affirm the design's efficacy and underscore its potential as a versatile and economically viable solution for aerial applications.
\end{abstract}

\section{Introduction} 



The concept of minimalist aerial vehicles is a novel area in aerial robotics. The popularity of quadrotors, primarily comprising four motors, an IMU, and a basic flight controller, has highlighted the need for more efficient designs. Their energy efficiency, while adequate, can be improved. It has been shown that despite the loss of up to three propellers, controlled flight is still possible\cite{mueller2014stability}. The challenge is to create aerial platforms that are highly functional yet minimize hardware complexity and simplify control algorithms. This approach, inspired by feasibility studies \cite{orsag_spincopter_2013} and culminating in practical models \cite{gress_using_2002}, offers significant benefits: streamlined control algorithms enhance reliability and ease of operation, while a reduced mechanical footprint decreases weight and power consumption, increasing efficiency and potential applications in diverse environments.


%
In the aerial robotics literature, there has been a concerted effort to reduce vehicle complexity. Bio-inspired designs have been at the forefront of this endeavor, leading to innovative approaches. 
For instance, the Samara wing uses a single wing and actuator and relies on cyclic control \cite{ulrich_control_2010}. Similarly, the foldable single actuator monocopter (F-SAM) employs a semi-rigid, foldable Samara wing \cite{win_design_2021}. 
Dual-actuation systems have origins in bicopter configurations \cite{gress_using_2002}\cite{cai_modeling_2023} but have also been influenced by natural auto-rotating behaviors, such as in dual-actuated spinning samaras \cite{bai_bioinspired_2022}\cite{9882300}\cite{orsag_spincopter_2013} and flapping-wing mechanisms \cite{ma_controlled_2013}. 
Further complexities like tricopters \cite{rongier_kinematic_2005, salazar-cruz_stabilization_2009} and quadrotors \cite{pounds_design_2002} offer additional controllable degrees of freedom, as they can hover without rotating. However, when it comes to tasks such as patrolling, random walking, flocking, and rendezvousing, it is worth questioning whether fewer controllable degrees of freedom could accomplish similar objectives \cite{5783895}. Research into continuous mapping with multirotors \cite{kaul_continuous-time_2016} and self-rotating vehicles \cite{chen_self-rotating_2023} opens avenues for identifying inherent solutions or
behaviors that can perform the required tasks with less complexity.
The proposed vehicle has the potential to be an aid in applications such as patrolling~\cite{WANG2020103402}, localization~\cite{muller2013efficient}, air and turbulence monitoring ~\cite{tan2012twin, arun2006atmos}.

\begin{figure}[t]
    \centering
    \includegraphics[, trim=2cm 2cm 2cm 2cm,clip, width=.85\linewidth]{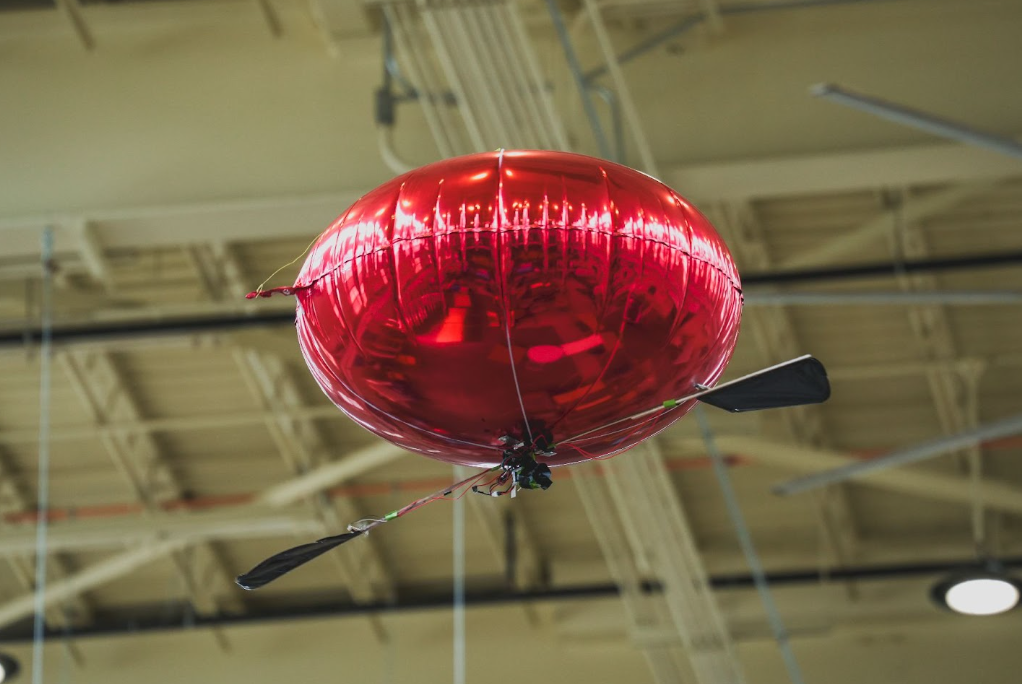}
    \caption{Spinning Blimp in action.}
    \label{fig:mainrobot}
\end{figure}

Existing spinning multi-rotor vehicles, as discussed in the literature, face two primary challenges. Firstly, they struggle to maintain stable attitude control without high-precision sensors or when the system is unbalanced. Secondly, heavier vehicles of similar design require higher angular velocities and more powerful motors to stay aloft, resulting in short flight durations (e.g., 24.5 minutes for the Revolving-wing Drone\cite{gress_using_2002}). To address these limitations, we propose an LTA (lighter-than-air) vehicle (see Fig.~\ref{fig:mainrobot}) that leverages the natural buoyancy force to remain airborne with minimal energy consumption. Furthermore, we incorporate inherent pendulum-like attitude stabilization allowing the omission of a complex stability control algorithm~\cite{xu2023sblimp}. 


The contributions of this paper are twofold. First, we propose a simplified control technique that significantly reduces the number of tuning parameters. Our controller proposes a cyclic controller with bang-bang control for accurate position tracking \cite{seyde2021bangbang}. Second, we present a novel aerial robot design that leverages spinning dynamics and pendulum-like stability to offer a low-cost, highly capable and safe solution for aerial applications.

\section{Vehicle Design} \label{sec:design}
\begin{figure}[t]
    \centering
    \includegraphics[width=.98\linewidth]{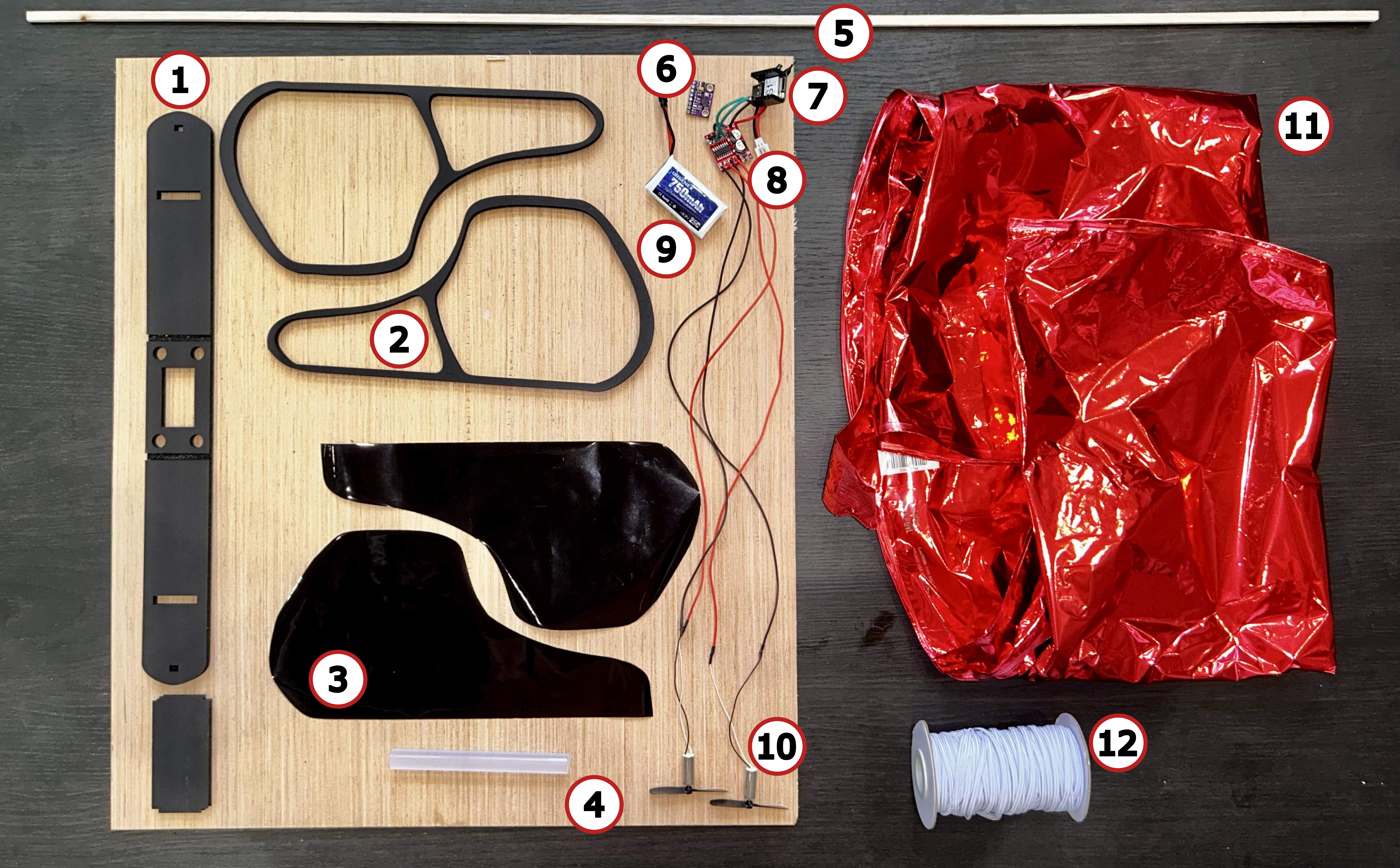}
    \caption{Components: 1) chassis, 2) foam core wing rib, 3) cellophane, 4) hot glue stick, 5) balsa wood s, 6) IMU, 7) flight computer, 8) DC motor driver, 9) LiPo Battery, 10) DC brushed motor, 11) mylar balloon, and 12) elastic string.}
    \label{fig:flatpack}
\end{figure}







The design of Spinning Blimp is focused on creating a minimalist aerial vehicle platform that can \revisions{hover or} \qbreview{R2 Q1} translate in midair for long periods of time while spinning. 
The center of mass of the vehicle is on the bottom of the balloon, creating a pendulum-like behavior that naturally maintains its horizontal attitude passively.
Two motors push two wings in a propeller-like configuration to generate a spinning behavior.

Our design is based on the following components (illustrated in Fig.~\ref{fig:flatpack}):

\paragraph{Helium Balloon} The balloon is made from Mylar and has an oblate spheroid shape.
    The total volume of the balloon is 0.081 $m^3$ and is filled with helium.
    The helium-filled balloon provides a positive buoyancy of 60 g. The net buoyancy of the vehicle, including the actuators and electronics, is -5 g.

\paragraph{Frame}
The frame is made from Elmer's foam core that weighs 224 g per 30$\times$20 sq inch sheet. Coming from a flat sheet, it is laser-cut and folded to form a cage.
The frame is attached to the balloon using two elastic strings. 
Passing through the frame, a square balsa wood stick forms a rigid axle supporting the wings and motors.
We designed and evaluated two vehicles with axles of 0.7 m and 1.3 m. 
The axle length creates effects that have been deeply studied in Blade Element Momentum Theory (BEMT) for a propeller \cite{driessens_triangular_2015,sammut_australasian_2005}.


\paragraph{Wings} 
The wings are made of foam core and laser-cut (see Fig.~\ref{fig:flatpack}).
 The wing design is made from cellophane and laser-cut foam core. The low angular velocity of the system presents two advantages. Lightweight and minimal strengthening of wings is required and high angles of attack are afforded by the intermediate flow regime ($Re < 10^{4})$. The design chosen in this paper focuses on producing a wing design with sufficient lifting capacity and rigidity for use in swarms where bumping behaviour is present and thus characterisation/evaluation of wing design parameter is not a primary focus of the study.
Wing configuration is investigated in two methods: \textit{i)} angle of attack from relative airflow and \textit{ii)} position the arm from the axis of rotation. Combined configurations for translational control are investigated based on the angular velocity and rate of the procession from the dis-symmetry of lift.

\paragraph{Motors}
The propulsion system is based on two brushed DC motors placed on each end of the support axle.
Each motor, in combination with a plastic propeller, can generate up to 15 g of static thrust.

\begin{figure}[t]
    \centering
     \includegraphics[width=.98\linewidth]{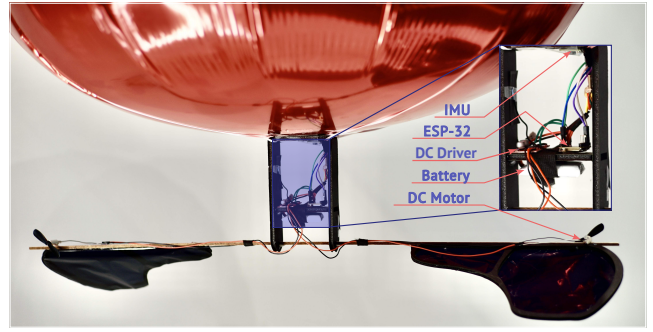}
      \caption{Electronics mounted on the Spinning Blimp.}
       \label{fig:diagram}
\end{figure}

\paragraph{Flight controller}
The hardware controlling the Spinning Blimp is a XIAO ESP32C3, a small 32-bit microprocessor supporting Wi-Fi communication protocols. This flight controller allowed us to iterate quickly using the Arduino IDE and choose from a variety of supported sensors through I2C, whether it is a GY-91 10-degree-of-freedom sensor or a BNO055 sensor.
A PWM signal, from the ESP32 flight controller, is sent to the DC brushed motors using a WWZMDiB L298N 2-channel DC motor driver. The whole system is powered via a one cell 3.7 V LiPo battery (750 mAh capacity).
The flight controller and all the electronics, except the motors, are mounted on the frame, as shown in Fig.~\ref{fig:diagram}.


\paragraph{Sensors}
In indoor experiments, we used the motion capture system (MoCAP), and no on-board sensors were needed.
For experiments without MoCAP, Spinning Blimp relies on yaw feedback data. This is achieved with the BNO055 sensor. Through sensor fusion of a gyroscope, accelerometer and magnetometer, Spinning Blimp is able to control its translational motion. Altitude control is achieved using a BMP280 barometric pressure sensor for height feedback. Random walk behaviors utilize a VL53L1X time of flight (ToF) sensor for wall detection.

\noindent
\textbf{Low-cost robot:}
One of the main advantages of this vehicle is its low cost, totaling around \$20 USD.
The vehicle uses minimal yet effective components: the XIAO ESP32C3 flight computer (\$4.99), GY-91 IMU (\$4.70), two DC motors (\$1.65), a DC driver (\$1.49), and two Gemfan 45mm propellers (\$0.74). The URGENEX 750mAh battery costs \$3.80, 10mm elastic bands (\$0.50), while the structure consists of 5mm foam core (\$1.75), and a Mylar balloon (\$0.78). Helium for lift costs \$0.24, making the vehicle a highly affordable option for aerial robotics experiments.

\section{Model}

\begin{figure}[t]
    \centering
     \includegraphics[width=1\linewidth]{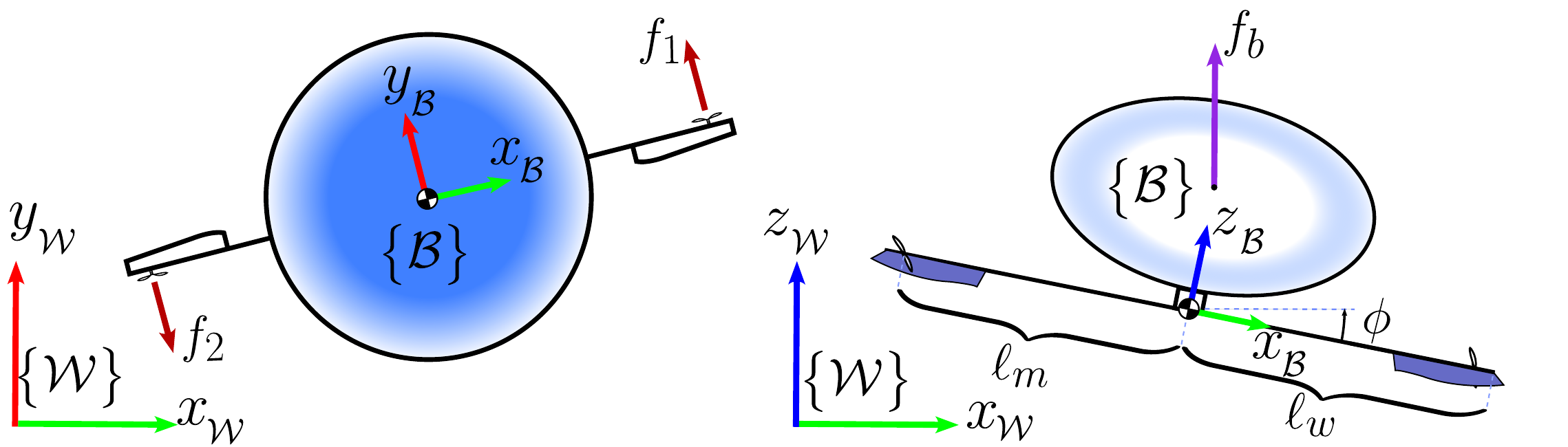}
      \caption{ 
      Diagram of coordinate reference frames in the world reference frame and body reference frame attached to the center of gravity (CG) of Spinning Blimp (left) and the relevant forces contributing to the dynamics of the system  (right).
      }
      \vspace{-1em}
       \label{fig:coordinates}
\end{figure}


We model the vehicle as a rigid body with mass $m$ that can move in the three-dimensional space $\mathbb{R}^3$.
The world frame, denoted by $\mathcal{W}$, is a fixed reference frame whose $z$-axis points upwards.
The body frame $\mathcal{B}$ is located on the frame, in the middle of the wing axle, with the $x$-axis 
pointing towards the right side of the axle and the $z$-axis pointing upwards towards the balloon. 
The robot's position in the world frame is denoted by $\boldsymbol{x}\in\mathbb{R}^3$, and its attitude is denoted by roll $\phi$, pitch $\theta$, and yaw $\psi$.
Its angular velocity in the body frame is $\boldsymbol{\omega}=[\omega_x, \omega_y, \omega_z]^\top\in\mathbb{R}^3$.

The \revisions{motors' rotation axes are parallel} \qbreview{R2 Q2}
to the $xy$-plane of the body frame.
Each motor $i=1,2$ generates a force $f_i\geq 0$, located at $\boldsymbol p_1={[\revisions{\ell_m},0,0]}$ and $\boldsymbol p_2={[-\revisions{\ell_m},0,0]}$ \qbreview{R2 Q3} in $\mathcal{B}$.
The spinning motion around the $z$-axis of the body frame 
drags the air using the wings as a propeller, generating a force with magnitude $k_{l} \omega_{z}^2$, where $k_{l}$ is a coefficient that abstracts wing parameters such as weight, surface area, and angle of attack.


\subsection{General dynamics}


    

We use Newton's equations to describe the translational motion of the vehicle.
The linear acceleration $\ddot{\boldsymbol{x}}$ is related to the 
vector forces generated by the motors $\boldsymbol{f}_{m}$,
balloon buoyancy and gravity $\boldsymbol{f}_{bg}$,
lift from the wings $\boldsymbol{f}_{l}$,
and air drag on the vehicle $\boldsymbol{f}_{d}$. Resulting into        
\begin{equation} 
        m
        \ddot{\boldsymbol{x}}
        = \boldsymbol{f}_{m} +
        \boldsymbol{f}_{bg} + 
        \boldsymbol{f}_{l} + 
        \boldsymbol{f}_{d}.
        \label{eq:newton1}
\end{equation}
The motor force $f_1$ points in the direction of the $y$-axis of $\mathcal{B}$ and $f_2$ points in the opposite direction (see Fig. \ref{fig:coordinates}). Let $\boldsymbol{u}=[f_1, f_2]^\top$ be the control input. Then, the force vector in the world frame is
$$
\boldsymbol{f}_{m} =
{}^W\!\!\boldsymbol{R}\!_B 
        \boldsymbol{A}\boldsymbol{u},
$$
where $^W\!\!\boldsymbol{R}\!_B\!\in$ SO(3) denotes the $3\times 3$ transformation matrix from the body frame to the world frame, 
$$
\boldsymbol{A} =
\begin{bmatrix}
        0&0\\ 1&-1\\ 0 &0\end{bmatrix}.
$$
The buoyant and the gravitational forces are in the inertial reference frame, and they can be combined as
$\boldsymbol{f}_{bg}= [0,0,f_{b}-mg ]^\top$, 
\revisions{where \(f_b\) is the upward force provided by helium, \(m\) is the total mass of the robot, and \(g\) is the gravitational acceleration}\qbreview{R1Q1b}, assuming negative buoyancy i.e.~$f_b - mg < 0$.

The spinning motion generates a lift force in the body frame that is transformed to the world frame,
$$
\boldsymbol{f}_{l} =  {}^W\!\!\boldsymbol{R}\!_B
       [0,0, k_{l} \, {\omega_z^2}]^\top.
$$
The air drag on the vehicle is assumed to be proportional to the square of the linear velocity of the vehicle.
$$
\boldsymbol{f}_{d}=
 {}^W\!\!\boldsymbol{R}\!_B 
 \begin{bmatrix}
       d_x\, \dot x  \,|\dot x|
       & d_y\, \dot y \,|\dot y |
       &d_z\, \dot z \,|\dot z|\\
        \end{bmatrix}^\top,
$$
where $(d_x,d_y,d_z)$ are drag coefficients.



The rotational dynamics is described using the Euler equation. 
The vehicle's inertia tensor is denoted by $\boldsymbol{I}$.
The torques on the vehicle are generated by the motors $\boldsymbol\tau_{m}$, buoyant force and gravity $\boldsymbol\tau_{bg}$,
and air drag $\boldsymbol\tau_{d}$. Then, 
\begin{equation}
\boldsymbol{I} \boldsymbol{\dot{\omega}} + \omega \times \boldsymbol{I}\omega = \boldsymbol\tau_{m} + \boldsymbol\tau_{bg} + \boldsymbol\tau_{d}, \text{ where}
\label{eq:euler}
\end{equation}
%
$$
\boldsymbol\tau_{m} = \sum_{i=1}^{2} \boldsymbol p_{i} \times 
\boldsymbol{A}\boldsymbol{u}
=
\ell_{m}\boldsymbol{B}\boldsymbol{u},
\text{ and }
\boldsymbol{B} =
\begin{bmatrix}
        0&0\\ 0&0\\ 1&1\end{bmatrix}.
$$
The gravitational and buoyant forces generate torques at the center of pressure $\boldsymbol{p}_{b}$ and center of mass $\boldsymbol{p}_{g}$ respectively,
$$
\boldsymbol\tau_{bg}= \boldsymbol p_{b} \times {}^B\!\!\boldsymbol{R}\!_W [0,0,f_b]^\top+ \boldsymbol p_{g} \times {}^B\!\!\boldsymbol{R}\!_W [0,0,-mg]^\top.
$$
The air drag is
$$
\boldsymbol\tau_{d} =[0,0,-k_w \omega_z^2]^\top.
$$
The combination of the buoyant force and the gravity force forms a system that acts like a pendulum.
So, we can assume that the  spinning blimp tends to remain horizontal with respect to the inertial frame 
to simplify the dynamics
\cite{xu2023sblimp}. 

\subsection{Simplified dynamics}
Assuming $\theta, \phi \approx 0$ for a horizontal robot, the transformation matrix $^W\!\!\boldsymbol{R}\!_B$ becomes a function of the angle yaw $\psi$ around the $z$-axis of the world frame, i.e, $^W\!\!\boldsymbol{R}\!_B\approx \boldsymbol{R}_z(\psi)$.
Then, the Newton's dynamics in \eqref{eq:newton1} is reduced to,
{
\begin{equation}
\label{eq:newton_simp}
            m
            \ddot{\boldsymbol{x}}
            =
            \!\boldsymbol{R}\!_z(\psi)          
           \boldsymbol{A}\boldsymbol{u}
            + 
            \begin{bmatrix}
            0\\ 0\\ f_{b}-mg \end{bmatrix}
            \\
            +
            \begin{bmatrix}
            0\\ 0\\ k_{lift} {\omega_z^2} \end{bmatrix}
            -             
        \begin{bmatrix}
       d_x\, \dot x  \,|\dot x|\\ d_y\, \dot y \,|\dot y |\\ d_z\, \dot z \,|\dot z|\\
        \end{bmatrix}.
\end{equation}
}
Assuming the vehicle is symmetric, the inertia tensor $ \boldsymbol{I}$ is diagonal.
The torque $\boldsymbol{\tau}_{bg}$ is zero when the vehicle is horizontal.
Then, the Euler's equation in \eqref{eq:euler} is reduced to


\begin{equation}
 \boldsymbol{I}\boldsymbol{\dot{\omega}} = 
 \ell_{m}\boldsymbol{B}\boldsymbol{u}+
 \begin{bmatrix} 0 & 0 & -d_{w}   \omega_z^2 \end{bmatrix}^\top.
 \label{eq:euler_simp}
\end{equation}
\section{Position Control}
\begin{figure}[b]
    \centering
    \includegraphics[width=0.9\linewidth]{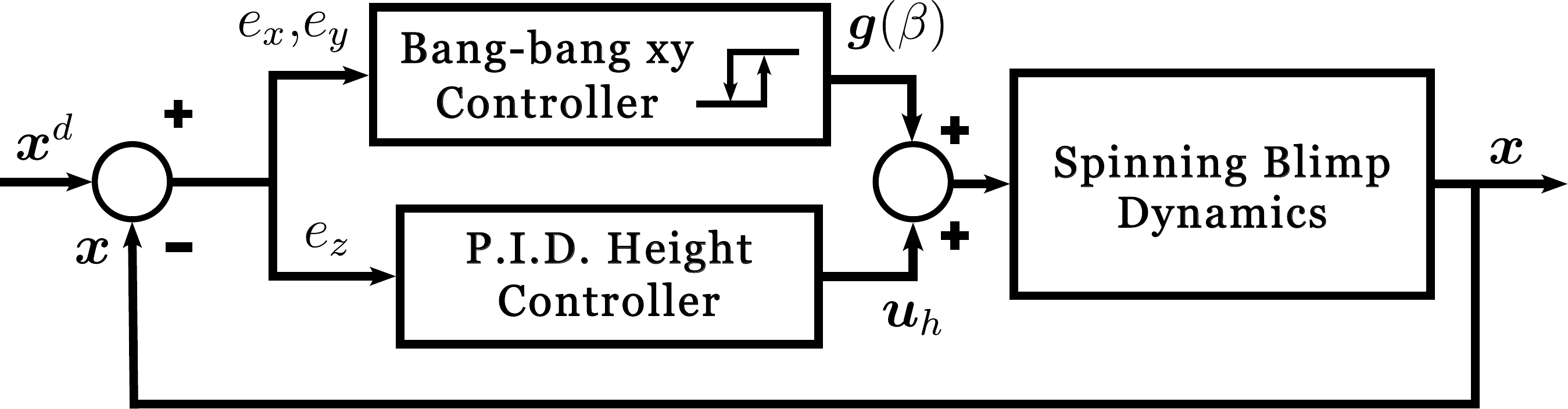}
    \caption{Control Block Diagram for position control.}
    \label{fig:controlloop}
\end{figure}
The goal of our position control is to move from a position $\boldsymbol{x}=[x,y,z]$ to a desired position $\boldsymbol{x}^d$.
We control the angular velocity $\omega_z$ to maintain a desired height $z^d$ while moving towards the goal on the $xy$-plane (see Fig.~\ref{fig:controlloop}).

\subsection{Height control}
To control the vehicle's height, we focus on the angular velocity $\omega_z$.
Looking at the $z$-axis of $\mathcal{W}$ in the Newton's equation \eqref{eq:newton_simp} when both motors generate the same force, we get

\begin{equation}
m \ddot z = f_b -mg - d_z\, \dot z \,|\dot z| + k_{lift} \, \omega_z^2.
\label{eq:z_dynamics}
\end{equation}
\revisions{The constant $k_{lift}$ is determined experimentally 
  by finding the angular velocity that makes the robot maintain a constant altitude. A systematic way to find the constant is using the bisection method~\cite{burden19852}.}
\qbreview{R1 Q1a and R2 Q4 how did we obtained klift? bisection}
Then, the height control can be achieved using
\begin{equation}
\omega_z^{*2} =\frac{-f_b+mg +K_p (z^d-z) + K_d (\dot z^d-\dot z) + \ddot z^d}{k_{lift}}.
\label{eq:controlz}
\end{equation}
The translational dynamics of the vehicle in the $z$-axis is independent of the $x$ and $y$-axes as it only depends on $\omega_z$. 
The control policy in \eqref{eq:controlz} leads the system to an asymptotically stable configuration. 
At constant angular velocities, aerodynamic damping influences the stability. 
%
We can achieve the desired angular velocity $\omega_z^*$ in \eqref{eq:controlz} using our control input in the dynamics of the the angular velocity around the z-axis described in \eqref{eq:euler_simp},
\begin{equation}
\boldsymbol{u}_h=\left (k\,(\omega_z^*-\omega_z)\,  
+ \frac{d_w}{2\ell_w} \omega_z^2 \, \right )
\begin{bmatrix}
    1 & 1
\end{bmatrix}^\top,
\label{eq:controlinputu}
\end{equation}
where $k>0$ is a constant gain.

\subsubsection{Stability in \(\omega_{z}\)}
By equating \eqref{eq:euler_simp} to \eqref{eq:controlinputu}, we obtain 
\begin{equation}
    \dot{\omega}_{z} = \frac{2kl}{I_{z}}(\omega_z^{*}-\omega_z).
\end{equation} 
This first-order differential equation has an exponential solution with
\(\omega \rightarrow \omega_z^{*}\) when time goes to infinity. This provides the necessary condition for the assumption that $\ddot{\psi} = 0$ and $\dot{\psi} = \omega_z^{*}$ and stability in z-axis.

\subsection{Motion on the plane}
The dynamics highlight how Spinning Blimp's position control is related to its heading.
To move on the $xy$-plane, the robot maintains the constant angular velocity in \eqref{eq:controlz} while creating a difference in the force between the motors. The effect of the difference is the translational motion that depends on the yaw angle $\psi$, as described in Newton's dynamics in \eqref{eq:newton_simp}.
 Cyclic or periodic motor forces allow for incremental contribution of motor forces towards a desired goal.
To achieve this goal, we will use a switching controller that will increase the thrust of the propeller in the direction of the goal while spinning.
  In order to maintain the constant angular velocity, that increase will be compensated by an equal reduction on the thrust on the other motor.
Our switching control strategy is simple but still produces similar results as sinusoidal cyclic control methods \cite{gress_using_2002} that require higher accuracy sensors and actuators.


\subsubsection{Bang-bang Control}



Bang-bang control, also denoted as on-off control or switching control, is a rudimentary control scheme that operates by activating or deactivating a control element in response to the system's measured state relative to a predefined threshold or setpoint\cite{seyde2021bangbang}. 
\begin{figure}[t] 
    \centering
    \begin{subfigure}[b]{0.54\linewidth}
        \centering
        \includegraphics[width=\linewidth]{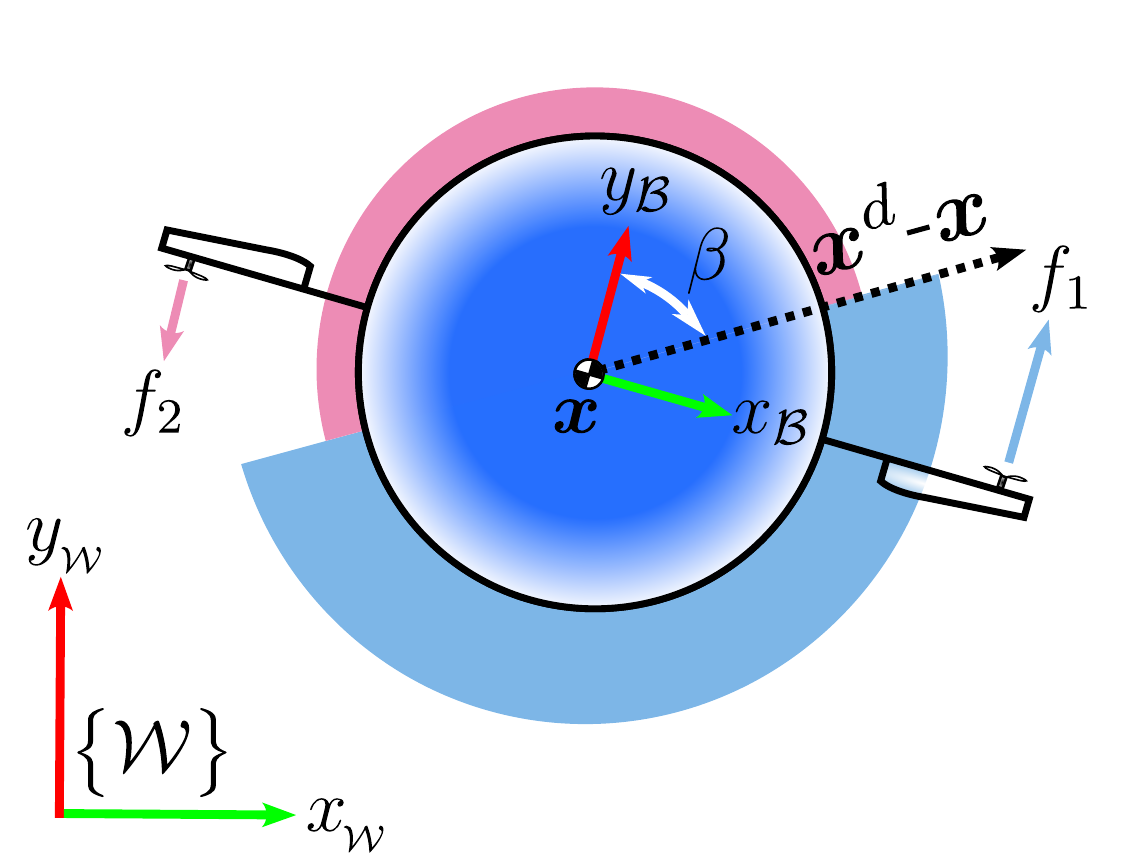}
        \caption{$-\frac{\pi}{2} < \beta < \frac{\pi}{2}$}
    \end{subfigure}
    \begin{subfigure}[b]{0.43 \linewidth}
        \centering
        \includegraphics[width=\linewidth]{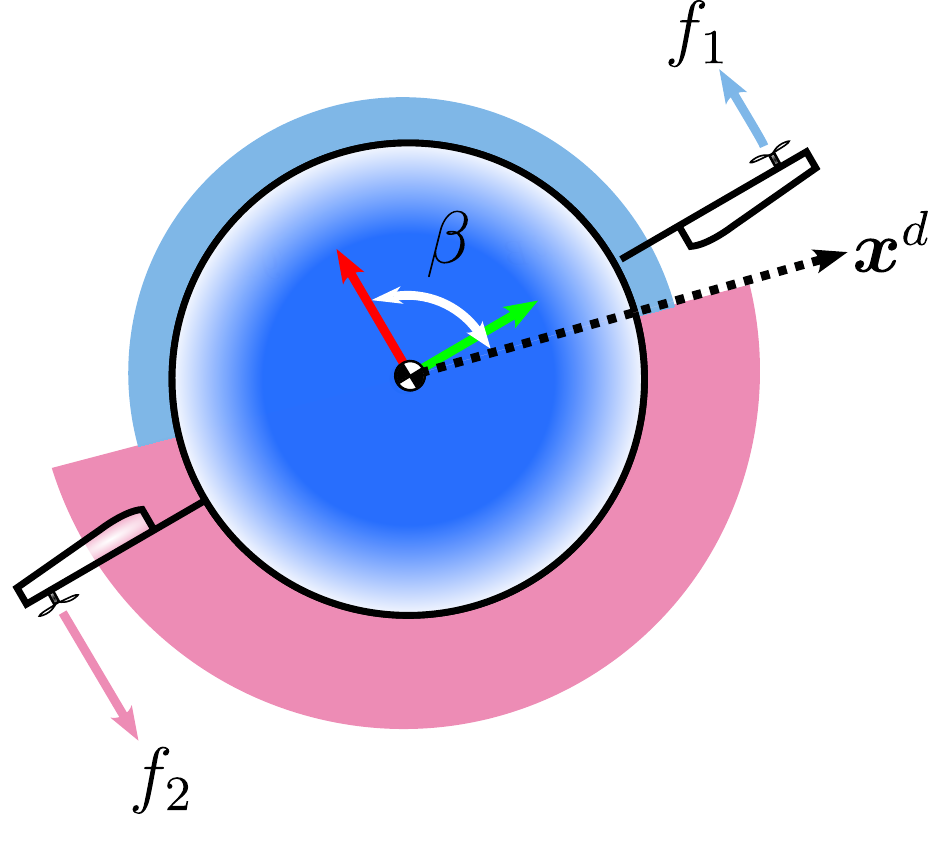}
        \caption{$\frac{\pi}{2} < \beta < -\frac{\pi}{2} (+ 2\pi)$}
    \end{subfigure}
    \caption{The two states of our control policy based on the angle to the goal $\beta$. The radius of the shaded region illustrates the magnitude of the motor for a given $\beta$ angle. }
    \vspace{-1em}
    \label{fig:heading}
\end{figure}

Let $\beta$ be the angle between the $y$-axis of $\mathcal{B}$ and the vector $\boldsymbol{x}^d-\boldsymbol{x}$, both vectors projected on the $xy$-plane of the world frame (see Fig.~\ref{fig:heading} for reference).
The controlled setpoint Spinning Blimp uses to operate in a switching state is when the 
angle $\beta$ is within a desired region. 
Based on the control input $\boldsymbol{u}= [f_{1},f_{2}]^\top$, we define our position controller as
\begin{equation}
\boldsymbol{u} = \boldsymbol{u}_{h} + \tau \,{g}(\beta) \begin{bmatrix}
        -1\\
         1\\
        \end{bmatrix},
\label{eq:control_input_u}
\end{equation}
where $\tau>0$ is a constant value. We create a thrust difference between motors for translation toward the goal based on the switching function,
\begin{eqnarray}
    {g}
        (\beta) = \begin{cases}
        1
        \text{, if $-\frac{\pi}{2} < \beta \leq \frac{\pi}{2}$},\\        
       -1
        \text{, otherwise.}\\
    \end{cases}
    \label{eqn12}
\end{eqnarray}
        
In summary, our control policy combines the height control of the vehicle using its spinning motion, and a bang-bang controller that activates the propeller in the direction of the goal (see Fig. \ref{fig:controlloop}).


\subsection{Stability analysis}
We use Lyapunov stability to analyze the dynamical system. The Lyapunov conditions are as follows: \(V(0) = 0, \quad V(x) > 0\quad \forall x \neq 0, \quad \dot{V}(x) < 0 \quad \forall x \neq 0\).
To prove stability, we analyze stability along the $z$-axis independently of the $x$- and $y$-axes due to the chosen control strategy.

\subsubsection{Stability in the \(z\)-Axis}

First, let's consider the height dynamics described in \eqref{eq:z_dynamics}, the error $e_{z} = z^d - z$ and the control policy for the angular velocity \eqref{eq:controlz}. Where $\ddot{z}=0$ at constant angular velocity $\Bar{\omega}_z^2 = \frac{mg - f_b}{k_{lift}}$.

The Lyapunov function candidate for the \(z\)-axis is
\begin{equation}
V(e_{z}, \dot{e}_{z})= \frac{1}{2}m\, e_z^2 + \frac{1}{2} K_p \dot{e}_z^2,    
\label{eq:lyapunov}
\end{equation}
which is positive definite as long as $K_p>0$ and $m>0$.
Its time derivative is
\begin{equation}
\dot{V}(e_{z}, \dot{e}_{z}) = m\, \dot e_z \ddot{e}_z + K_p\, e_{z} \dot{e}_z.
\label{eq:lyapunov_derivative}
\end{equation}
The error dynamics  comes from substituting \eqref{eq:z_dynamics} with
\(\omega_z~=~\omega_z^*\), and $\dot e_z = -\dot z$,
\begin{equation}
\ddot{e}_z = \frac{1}{m}\left(-K_p e_z - K_d \dot e_z - d_z \, \dot e_z \,|\dot e_z|\right).
\end{equation}
Substituting the error dynamics in \eqref{eq:lyapunov_derivative},

\begin{equation}
\dot{V}(e_z, \dot{e}_z) = - K_d \dot{e}_z^2 - d_z \dot{e}_z^2 |\dot{e}_z|.
\label{eq:lyapunov_derivative_final}
\end{equation}
Since both terms in \eqref{eq:lyapunov_derivative_final} are negative for all non-zero \(\dot{e}_z\), the function \(\dot{V}(e_z, \dot{e}_z)\) remains strictly negative, except when \(\dot{e}_z = 0\). This strict negativity ensures that the energy of the system decreases over time, leading the error dynamics to converge to zero. Consequently, the system is driven toward the equilibrium point, establishing asymptotic stability along the \(z\)-axis and guaranteeing that both the position and velocity errors decrease.







\subsubsection{Stability on the \(xy\)-plane}

To analyze the stability of the system, we will examine the dynamics on the $xy$-plane.

We use relevant portions of \(\boldsymbol{R}_z(\psi)\) for $xy$-plane and define $$\boldsymbol{A}_{xy} = \begin{bmatrix} 
        0&0\\ 1&-1
\end{bmatrix},\text{ and error } \boldsymbol{e}_{xy} = \begin{bmatrix} 
        {x}^d - {x}\\ {y}^d - {y}
\end{bmatrix}.$$
 From  \eqref{eq:newton_simp}, the dynamics on the $xy-$plane is,
\begin{equation}
\label{eq:newton_xy}
            m
            \ddot{\boldsymbol{x}}_{xy}
            =
            \boldsymbol{R}_z(\psi)\boldsymbol{A}_{xy} \boldsymbol{u} - 
        \begin{bmatrix}
       d_x\, \dot x  \,|\dot x| & 
       d_y\, \dot y \,|\dot y |\\ 
        \end{bmatrix}^\top.
\end{equation}
By substituting the rotation matrix and control input, we obtain,

\begin{equation}
            m
            \ddot{\boldsymbol{x}}_{xy}
            =
           2\tau\, g(\beta) [-\sin \psi, \cos \psi]^\top-
        \begin{bmatrix}
       d_x\, \dot x  \,|\dot x| & 
       d_y\, \dot y \,|\dot y |\\ 
        \end{bmatrix}^\top.
\end{equation}
Without loss of generality, assuming the desired location is at the origin, the direction to the goal is $\beta = \psi+\pi$. 
Since  $\dot{\boldsymbol{x}}_{xy}= -\dot{\boldsymbol{e}}_{xy}$ and $\ddot{\boldsymbol{x}}_{xy}= -\ddot{\boldsymbol{e}}_{xy}$, the dynamics of the error is
\begin{equation*}
            \ddot{\boldsymbol{e}}_{xy}
            =
           - \frac{2\tau}{m} g(\beta) [\sin \psi, -\cos \psi]^\top 
           -
           \frac{1}{m}
        \begin{bmatrix}
       d_x\, \dot e_x  \,|\dot e_x| & 
       d_y\, \dot e_y \,|\dot e_y |\\ 
        \end{bmatrix}^\top.
\end{equation*}
\noindent
We define the Lyapunov function candidate as
\begin{equation}
V(\boldsymbol{e}_{xy}, \dot{\boldsymbol{e}}_{xy}) = \frac{1}{2} \boldsymbol{e}_{xy}^\top \boldsymbol{e}_{xy} + \frac{1}{2} \dot{\boldsymbol{e}}_{xy}^\top \dot{\boldsymbol{e}}_{xy},
\label{lyapunov_position_eqn}
\end{equation} 
which is positive definite.
Its time derivative is
\begin{equation}
\dot{V}(\boldsymbol{e}_{xy}, \dot{\boldsymbol{e}}_{xy}) = \boldsymbol{e}_{xy}^\top \dot{\boldsymbol{e}}_{xy} + \dot{\boldsymbol{e}}_{xy}^\top \ddot{\boldsymbol{e}}_{xy}. 
\label{lyapunov_position_derivative_eqn}
\end{equation}
Substituting the error dynamics, 

\begin{equation*}
\dot{V}(\boldsymbol{e}_{xy}, \dot{\boldsymbol{e}}_{xy}) = 
 \dot{\boldsymbol{e}}_{xy}^\top \Big(\boldsymbol{e}_{xy}
 -\frac{2\tau}{m}g(\beta)  
 \begin{bmatrix}
    \sin \psi\\
   - \cos \psi \\
 \end{bmatrix}\Big ) 
       - \frac{1}{m}\begin{bmatrix}
       d_x\, \dot e_x^2  \,|\dot e_x| \\ 
       d_y\, \dot e_y^2 \,|\dot e_y |\\ 
        \end{bmatrix}^\top    
\end{equation*}

\noindent
Because of the cubic positive damping term, \(\dot{\boldsymbol{e}}_{xy}\) does not grow indefinitely and remains in a finite region, implying that the system is uniformly ultimately bounded (UUB). Furthermore, since both \(g(\beta)\) and \(\psi\) are strictly increasing, the magnitude of 
\({\boldsymbol{e}}_{xy}-\frac{2\tau}{m}g(\beta) \begin{bmatrix} \sin \psi & - \cos \psi \end{bmatrix}^\top\)
decreases for large \(\|\boldsymbol{e}_{xy}\|\). However, when \(\|\boldsymbol{e}_{xy}\|\) is small, that term increases again, resulting in oscillatory behavior near the origin. 
\revisions{By LaSalle’s principle, trajectories converge to the largest invariant set in \(\{\dot{V}=0\}\). Damping keeps \(\dot{\boldsymbol{e}}_{xy}\) bounded, so \(\boldsymbol{e}_{xy}\) remains finite, and only small-limit oscillations arise. Thus, the error does not diverge, and the motion stays bounded near the origin.}

\section{Evaluation}
\begin{figure}[t]
    \centering
        \includegraphics[trim={0cm 0cm 0cm 0cm}, clip, width=.9\linewidth]{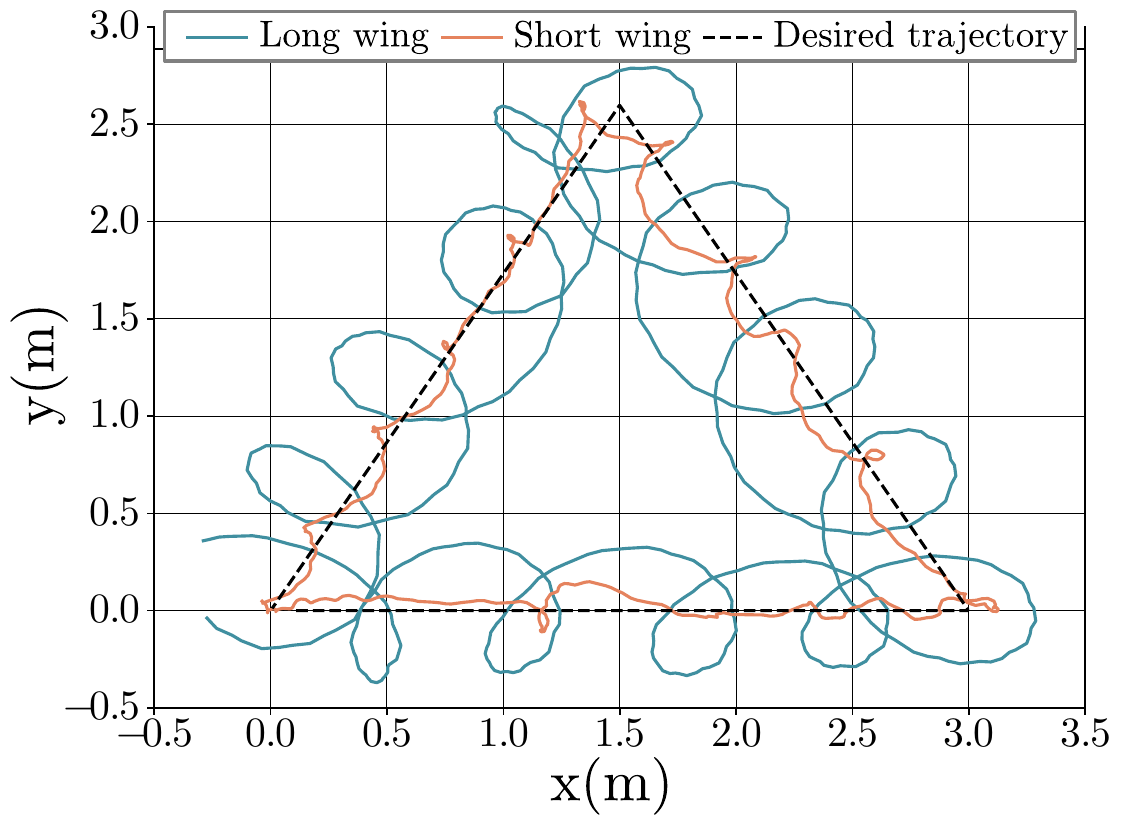}
    \caption{Plot in xy plane for counter clockwise triangle tracking ($v = $ 0.1 $\text{ms}^{-1}$)}
    \label{fig:trianglexy}
    \vspace{-1em}
\end{figure}
In our  experiments with actual robots,
we evaluate our Spinning Blimp while hovering for long periods of time, tracking trajectories with sharp turns and tracking non-simple trajectories. The evalution follows the convention for stability analysis using $\boldsymbol{e}=\|\boldsymbol{x}^d-\boldsymbol{x}\|$  (meters)\footnote{The source code for simulations and actual robots is available at\\ 
\href{https://github.com/LehighBlimpGroup/S_blimp-simulator}{https://github.com/spinning\_blimp}
}.
Furthermore, we will show a motivating experiment that illustrates a potential application of our robot.

%

\subsection{Experiment 1: Hover Endurance Test}
One of the most relevant design factors of the vehicle is the wing diameter ($2\ell_{w}$).
So, we will evaluate short-wings with 0.7m and long-wings with 1.3m.
Additionally, we also compare with other LTA vehicles such as the bi-copter \cite{xu2025mochiswarmtestbedroboticblimps, li2023novel}, and the S-Blimp \cite{xu2023sblimp}.
Our results show that the endurance for each vehicle is: 30mins for Bi-copter, 75 mins for S-Blimp, \textbf{65 mins} for Spinning Blimp - short-wing, and the robot with the most endurance is \textbf{78 min} for the Spinning Blimp with long-wings.

Even though the spinning blimp did not show a significant difference from other vehicles, such as the S-Blimp, 
it showed a long endurance and high stability during flight. 
Our robot does not require high-frequency sensors to operate, as it does not drift as much as other vehicles.
The spinning motion mitigates the imperfections in the actuators.

    
\begin{figure*}[t] 
    \centering
    \begin{subfigure}[b]{0.7\textwidth}
        \centering
        \includegraphics[width=\linewidth]{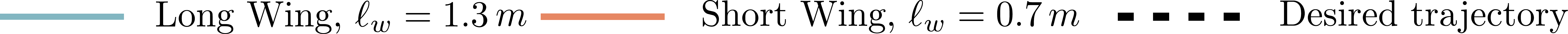}
    \end{subfigure}


    \begin{subfigure}[b]{0.3\textwidth}
        \centering
        \includegraphics[width=\linewidth]{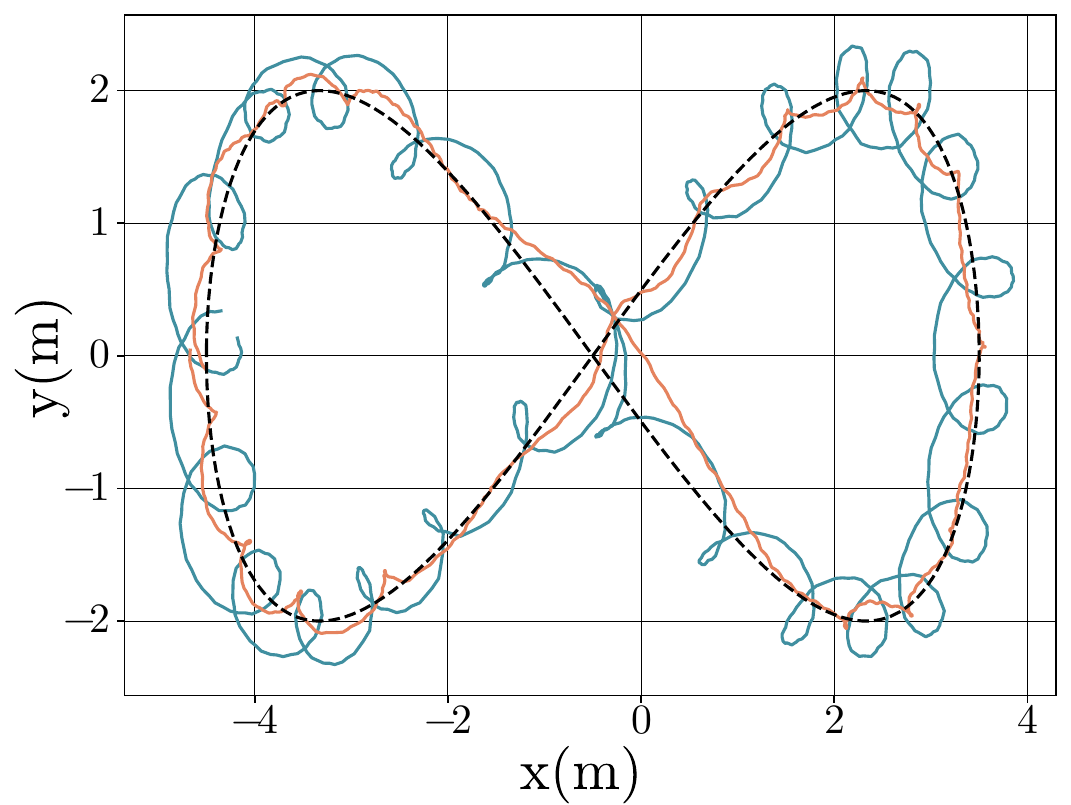}
        \caption{$v = 0.13 \text{ms}^{-1}$}
    \end{subfigure}
    \hfill
    \begin{subfigure}[b]{0.3\textwidth}
        \centering
        \includegraphics[width=\linewidth]{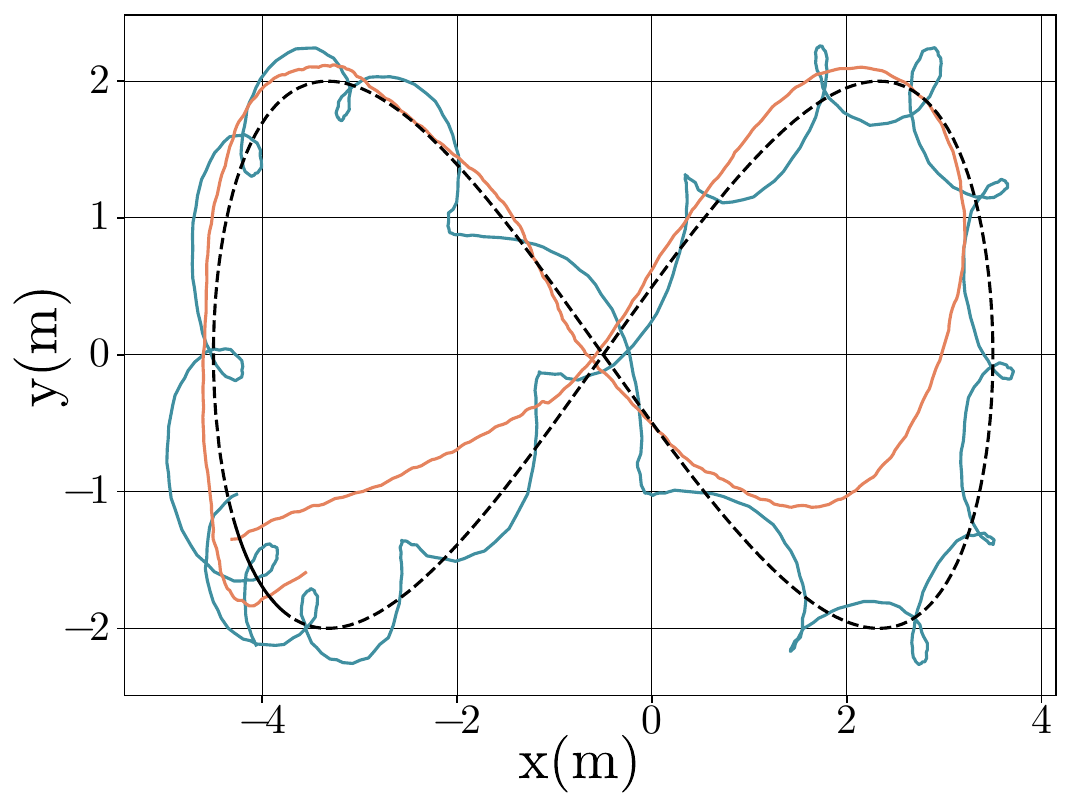}
        \caption{$v = 0.21 \text{ms}^{-1}$}
    \end{subfigure}
    \hfill
    \begin{subfigure}[b]{0.3\textwidth}
        \centering
        \includegraphics[width=\linewidth]{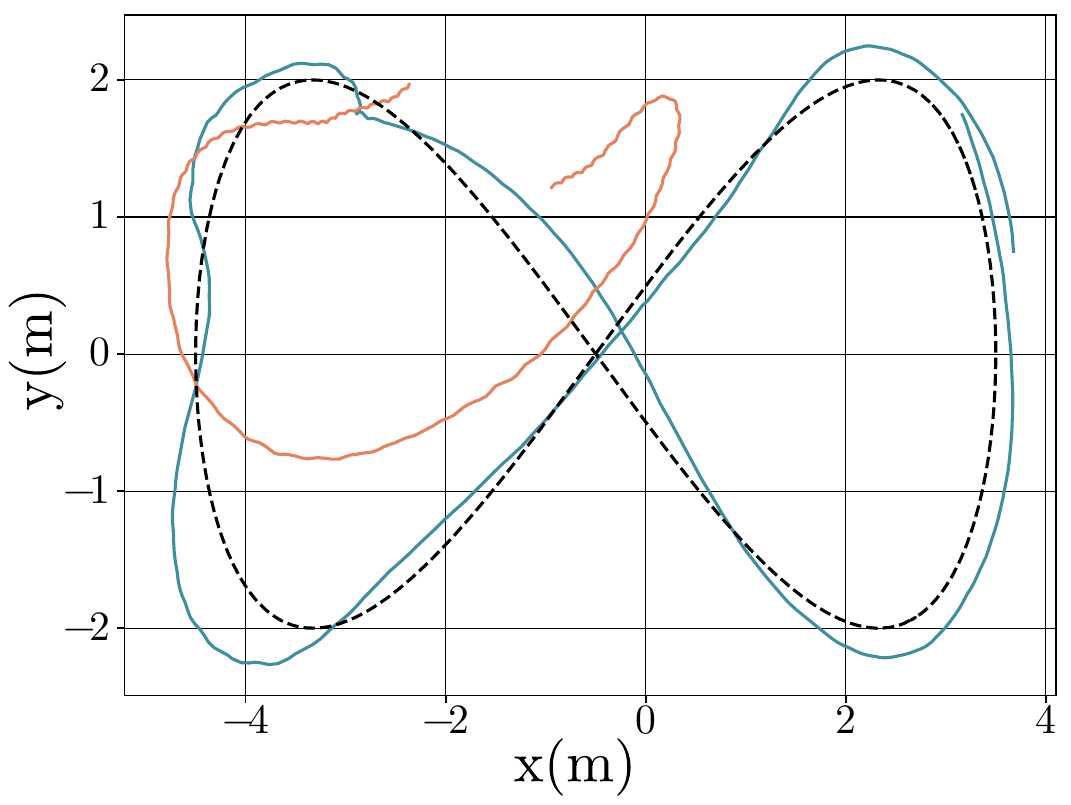}
        \caption{$v = 0.42 \text{ms}^{-1}$}
    \end{subfigure}

    \caption{Lissajous curve position tracking at three average velocities.}
    \vspace{-1em}
    \label{fig:lissajous}
\end{figure*}

\subsection{Experiment 2: Triangle Position Tracking}

We evaluate the ability to track a sharp turn by tracking an equilateral triangle (Fig.~\ref{fig:trianglexy}). 
We highlight that the direction of the trajectory goes in the opposite direction of the spinning motion. 
At a velocity of $v = $ 0.10 $\text{ms}^{-1}$ an equilateral triangle was tracked using our control policy.
Note, at the time of running the experiment, external disturbances from the air conditioning system in the lab showed a large impact on the short-wing Spinning Blimp. For the short-wing, during straight line trajectories and in the corners, overshoot is kept to a minimum. For the long-wing, minimal overshoot is present for all three corners, however, significant error is present from constantly overshooting the desired point.

\subsection{Experiment 3: Lissajous Position Tracking}

The non-linear and symmetrical trajectory chosen for Spinning Blimp is the Lissajous curve,
$$
\boldsymbol{x} = [A \sin(a(t) + \delta_x), B \sin(b(t) + \delta_y), z^d]^{\top}.
$$
 where $A=4, B=2, a=1, b=2, \delta_x= \pi/2, \delta_y=0$ are constants that augment the shape of the curve Fig.~\ref{fig:lissajous}.

The symmetrical nature of the Lissajous curve includes turns contributing to the direction of rotation and vice versa; allowing both controlled turns to present insights in one plot. For the short-wing Spinning Blimp, the average velocity of $v = 0.13\text{ms}^{-1}$ proved an upper bound for precise position tracking. Where consequent increases showed the limitation of a short lever arm for motor inputs and the absence of feed-forward control terms. Long-wing showed tracking capabilities at a wide range of velocities up to $v = 0.42 \text{ms}^{-1}$. 
Where short-wing presents strengths in remaining stable at tracking slow velocities and fast velocities with comparable mean Euclidean errors. 
Constant bang-bang $\tau$ values guarantee overshoot with long-wing Spinning Blimp. Overshooting the position highlights the periodic control input as a result of switching desired heading vector $\boldsymbol{x}^d-\boldsymbol{x}$.
As it can be seen in Fig.~\ref{fig:lissajous}, Spinning Blimp produces a behavioural tendency to overshoot to the outside of the curve in the clockwise portion and overshoot to the inside of the curve during the counter-clockwise portion.

\subsection{Experiment 4: Exploration through random walk}

The LTA nature and slow spinning of the vehicle make it safe for operation in environments that may include human interaction, making it a good choice for indoor exploration tasks.
The platform's stable flight allows for the addition of various sensory equipment, providing rich data collection capabilities essential for exploration. 

\begin{figure}[b]
    \centering
    \includegraphics[width=0.95\linewidth]{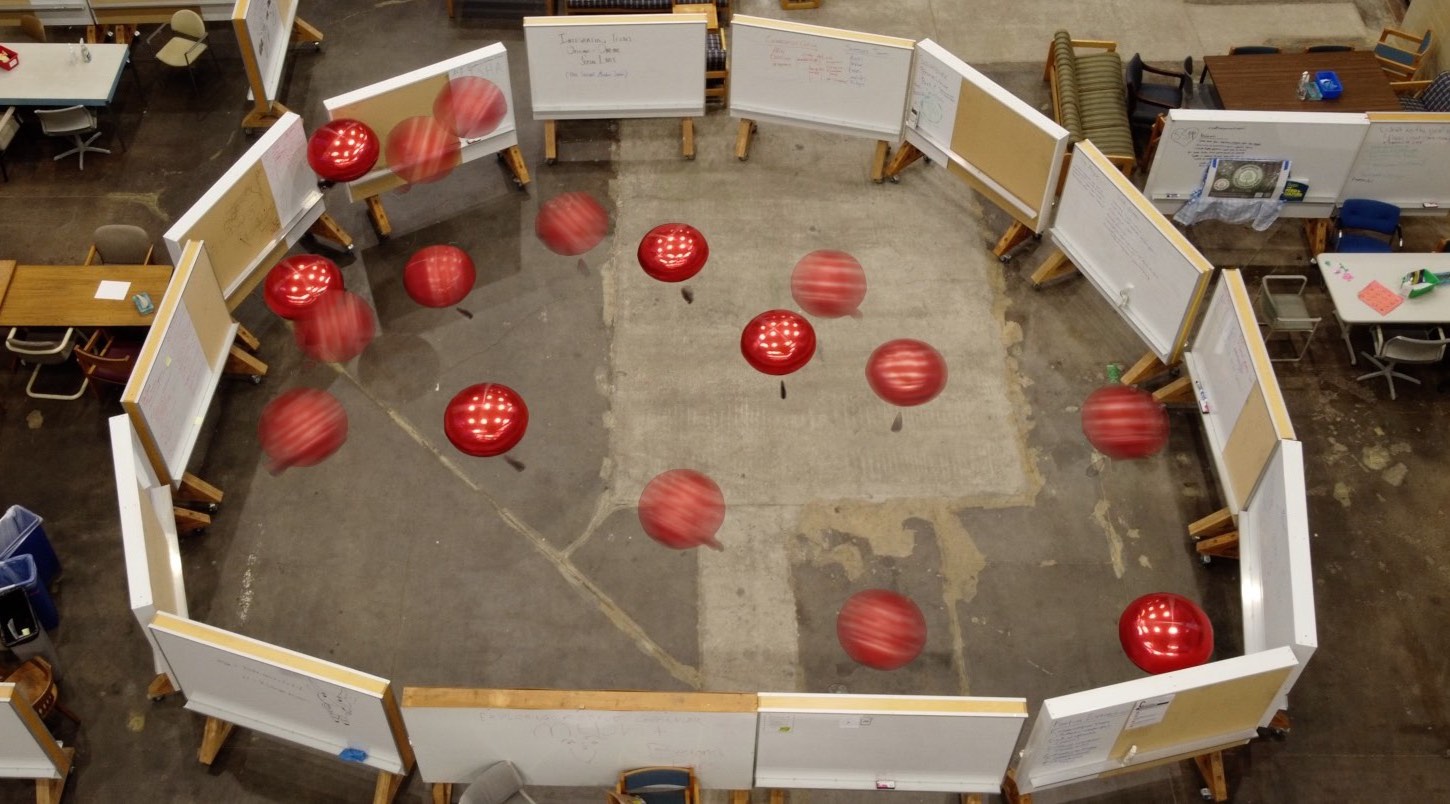}
    \caption{Random Walk and Collision Resistance Behaviour}
    \label{fig:randomwalk}
\end{figure}

In our experiment, we equipped the spinning blimp with a time of flight sensor (ToF) that can measure distances of up to 2.5 m.
The random walk behaviour lends itself well to a spinning vehicle. By rotating 360$^{\circ} \approx$ every 0.5 s, a ToF sensor can record distances around itself. Spinning blimp's natural stability combined with a fixed velocity in $x$ and $y$ create a valuable exploration platform. When approaching a wall Spinning Blimp halts the forward velocity within a threshold distance. Taking the normal vector, based on body frame yaw angle corresponding to the minimum distance recorded, a reflection direction for forward velocity is calculated. When it collided, neither the object struck nor Spinning Blimp sustained any damage; highlighting Spinning Blimp's suitability for indoor mapping tasks. After multiple trials and bouncing around the environment (Fig.~\ref{fig:randomwalk}), we show how the Spinning Blimp can be a safe and promising platform for exploring new environments.

\section{Discussion}
The use of brushed DC motors in our system is motivated by lower cost and weight savings. However, the robot also takes disadvantages from brushed DC motors such as lower accuracy and longer speed-up and slow-down time \cite{7837471}. Different actuators, e.g. brushless motors, would provide more responsive control at the cost of weight and subsequently higher inertial forces to overcome. The bang-bang control policy aims to alleviate motor ramp up concerns. Through small DC motors, Spinning Blimp presents a safe aerial platform for indoor research around humans. 


For robustness against wind disturbances, 
the short-wing version of Spinning Blimp showed minimal resilience to constant and gusting wind conditions. However, the long-wing span version showed resilience from a constant wind velocity up to $v = 0.5 - 1.5 \text{ms}^{-1}$. Combined with minimal impact to attitude, the long-wing design presents a platform more suitable to unpredictable environments.

\section{Conclusion and Future Work}
The Spinning Blimp presents an economical yet powerful platform. The combination of an LTA gas and continuously revolving dynamics produces inherent attitude stability, ensuring that forces contribute solely to translational movements. 
The vehicle's design and the accompanying simplified control algorithm have proven effective in achieving precise position tracking with reduced tuning complexity. Applications of the Spinning Blimp extend from patrolling and localization to environmental monitoring, making it a versatile and cost-effective solution for both research and practical purposes.
In our future work, we plan to explore swarm behaviors that include rendezvous, formation, and flocking of spinning vehicles.


\bibliographystyle{IEEEtran}
\bibliography{references}

\end{document}